\UseRawInputEncoding
\documentclass[pre,aps,twocolumn,amsmath,amssymb,longbibliography]{revtex4-1}

\pdfoutput=1
\usepackage{hyperref}
\usepackage{graphicx}
\usepackage[usenames]{color}
\usepackage{bm}
\usepackage[final]{movie15}

\definecolor{blue}{rgb}{0,0,0}

\def\cal#1{\mathcal{#1}}
\def\eqq#1{Eq.~(\ref{#1})}
\def\eq#1{(\ref{#1})}
\def\av#1{\langle #1 \rangle}

\def\f#1{Fig.~\ref{#1}}

\def\c#1{~\cite{#1}}
\def\cc#1{Ref.\c{#1}}

\def\ta0{\tilde{a}_0}

\def\s#1{Section~\ref{#1}}

\def\beq{\begin{equation}}
\def\eeq{\end{equation}}
\def\bea{\begin{eqnarray}}
\def\eea{\end{eqnarray}}

\begin{document}

\title{Cellular automata can classify data by inducing trajectory phase coexistence}
\author{Stephen Whitelam$^1$}
\email{{swhitelam@lbl.gov}} 
\author{Isaac Tamblyn$^{2,3}$}
\email{{isaac.tamblyn@uottawa.ca}}
\affiliation{$^1$Molecular Foundry, Lawrence Berkeley National Laboratory, 1 Cyclotron Road, Berkeley, CA 94720, USA\\
$^2$University of Ottawa, Ottawa, ON K1N 6N5, Canada\\ $^{3}$Vector Institute for Artificial Intelligence, Toronto, ON M5G 1M1, Canada}

\begin{abstract}
We show that cellular automata can classify data by inducing a form of dynamical phase coexistence. We use Monte Carlo methods to search for general two-dimensional deterministic automata that classify images on the basis of activity, the number of state changes that occur in a trajectory initiated from the image. When the number of timesteps of the automaton is a trainable parameter, the search scheme identifies automata that generate a population of dynamical trajectories displaying high or low activity, depending on initial conditions. Automata of this nature behave as nonlinear activation functions with an output that is effectively binary, resembling an emergent version of a spiking neuron.
\end{abstract}

\maketitle

\section{Introduction} 

Cellular automata are discrete dynamical systems. They are widely studied because they are simple to specify and display considerable complexity, including the ability to do universal computation\c{von2017general,conway1970game,wolfram1984universality,wolfram1983statistical,chopard1998cellular,sante2010cellular}. Cellular automata have been used to do computations such as classification, either directly\c{ganguly2002evolving,maji2003theory}, or as a reservoir that enacts a nonlinear transformation prior to additional calculation\c{yilmaz2015machine,nichele2017reservoir,moran2018reservoir}, or in combination with a neural network\c{randazzo2020self}. Cellular automata do not currently perform as well at most machine-learning tasks as deep neural networks, but continue to attract attention because discrete automata specified by integers consume less memory and power than do real-valued neural nets, a property useful for mobile devices\c{moran2019energy}.

In this paper we search for general two-dimensional cellular automata able to classify images in MNIST\c{lecun1995learning,mnist}, a standard data set. While image classification has been done by automata\c{randazzo2020self,moran2018reservoir}, our focus is {\rm how} automata learn to do classification, and we describe a mechanism of decision making that resembles the phase transitions undergone by physical systems such as magnets\c{kinzel1985phase,binney1992theory} or glasses\c{hedges2009dynamic}. 

This connection is enabled by the introduction of {\em activity} as an order parameter for classification. We define activity as the total number of changes of state of all cells within a dynamical trajectory. Used in the study of glasses\c{meshkov1997low,hedges2009dynamic}, it is a natural candidate for computations such as image classification because it is invariant to translations and is not biased toward particular spatial environments. If the number of timesteps of the automaton (we call this the {\em depth} of the automaton) is a trainable parameter, we show that Monte Carlo search of the automaton rule table identifies rules and depths that perform classification by enacting a form of dynamical phase coexistence. The action of these automata on the set of MNIST images produces a set of dynamical trajectories of two distinct types, possessing large or small activity. Automata of this nature resemble an emergent version of a spiking neuron\c{paugam2012computing,gerstner2002spiking}, enacting a temporal calculation before committing to a state of large or small output. We show that collections or reservoirs of such automata can be used by a linear model to classify images with an accuracy comparable to that of fully-connected neural networks that act on the pixels of the original image.

The automata described here perform classification by inducing a form of dynamical phase coexistence. Phase transitions and phase coexistence are seen in cellular automata in a variety of different settings. Probabilistic automata such as the Ising model capture the essence of phase transitions in physical systems such as fluids or magnets\c{kinzel1985phase,binney1992theory}. Probabilistic and deterministic automata modeling traffic flow show coexistence of spatial patterns\c{jiang2002cellular,wolf1999cellular,neto2011phase,d2005coexisting}, and exhibit dynamical phase transitions as a function of density\c{schadschneider1999new,yukawa1994dynamical,d2005coexisting}. There is also evidence of a sharp transition in the space of rules\c{li1990transition,wootters1990there} between different classes of deterministic automata\c{wolfram1984universality}. The phase coexistence seen here occurs for deterministic automata of finite spatial extent, for sufficiently long trajectories. The trajectories in question contain a variety of spatial patterns but are defined by their dynamics, which falls into one of two classes and is accompanied by a bimodal distribution of a time-integrated order parameter. This bimodality is similar to that seen in stochastic models of growth\c{klymko2017similarity} or the conditioned (atypical) trajectories of atomistic glasses\c{hedges2009dynamic}. In this case the coexistence emerges in response to an instruction to do classification. It occurs in a many-body system in the absence of explicit feedback, and in the typical (indeed deterministic) trajectory ensemble.  

In \s{model} we introduce the model and simulation methods. In \s{pc} we show that a cellular automaton can classify images by enacting a form of dynamical phase coexistence. In \s{acc} we discuss the accuracy with which linear models acting on a collection of automata can classify images. We conclude in \s{conc}.

\section{Model and simulation details} 
\label{model}

We consider the MNIST data set, a collection of $28 \times 28$-pixel grayscale images of 70,000 handwritten digits of class $\alpha = 0,1,\dots,9$\c{lecun1995learning,mnist}. We binarized images by setting to 0 or 1 any pixel whose value is less than or greater than $1/4$ of the maximum possible intensity, respectively. One of these images is shown in \f{fig_schematic}(a). These binary images serve as the initial conditions for simulations, which take place on a square lattice of size $N=28 \times 28$ with periodic boundary conditions. On each site $i \in[1,N]$ of the lattice lives a binary cell $S_i=0,1$, indicated white (0) or blue (1) in images. We evolve the lattice in time using a discrete-time deterministic cellular automaton. The automaton is a table of rules $S_i(t+1) = {\cal R}(s_i(t))$ that specifies the state of cell $i$ at time $t+1$ given the state $s_i(t)$ of its environment $\{S_i(t)\}$ at time $t$. The table ${\cal R}(s)$, where $s\in[0,511]$, consists of 512 integers 0 or 1, and is defined as follows.

The environment of cell $i$ is the nine-neighbor-square (Moore) construction\c{packard1985two} shown in \f{fig_schematic}(b), denoted $\{S_i\}=(S_i^{(0)},S_i^{(1)},\dots, S_i^{(8)})$. Cells in this environment are indicated $S_i^{(m)}$, the $m^{\rm th}$ neighbor of the central site $i$, where $m=0,1,2,\dots 8$. As indicated in the figure, we define $m=0$ as the bottom-left site, and increment $m$ by moving left to right, row by row (note that $S_i^{(4)}=S_i$). The state of the environment of cell $i$ at time $t$ is indicated by the integer 
\beq
\label{enviro}
s_i(t)=\sum_{m=0}^8 2^m S_i^{(m)}(t),
\eeq
 where $S_i^{(m)}(t) =0,1$ is the spin state of neighbor $m$ at time $t$. For example, the state of the environment with all cells zero (white) is $s_i=0$, and the state of the environment with all cells unity (blue) is $s_i=2^9-1=511$. The state of the environment shown in \f{fig_schematic}(b) is $s_i=2^1+2^3+2^5+2^7=170$. 
 \begin{figure}[] 
   \centering
  \includegraphics[width=\linewidth]{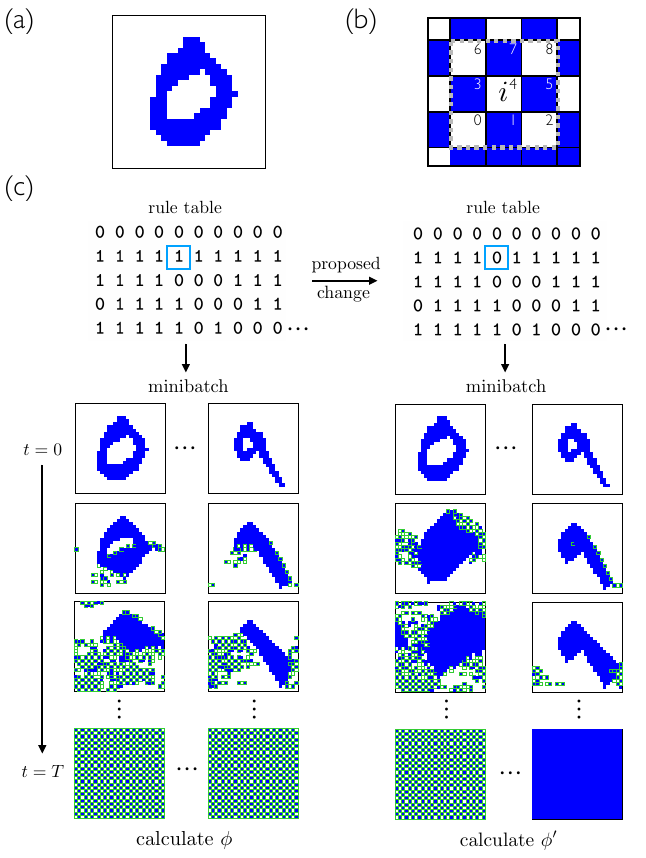} 
   \caption{(a) Example binarized MNIST digit on a $28 \times 28$ lattice. (b) Example of a $3 \times 3$ Moore neighborhood of site $i$, defining the 512-entry rule table of the cellular automaton. (c) Schematic of the Monte Carlo search procedure on the automaton rule table, designed to maximize $\phi$, \eqq{phi}. Blue and white cells indicate states 1 and 0; green borders indicate cells whose states have changed from the previous time step.}
   \label{fig_schematic}
\end{figure}

\begin{figure}[] 
   \centering
  \includegraphics[width=\linewidth]{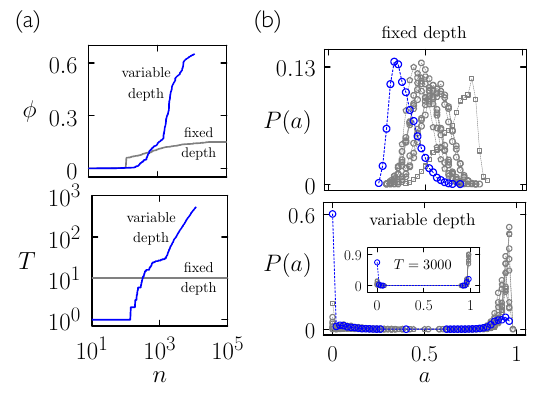} 
   \caption{(a) Monte Carlo search of the automaton rule table to maximize $\phi$, \eqq{phi}, for digit class $\alpha=0$. The two panels show $\phi$ and automaton depth $T$ as a function of Monte Carlo step number $n$, for simulations carried out in fixed-depth mode ($T=10$; gray) and variable-depth mode (blue).  (b) Histograms of activity, \eqq{ay}, for 60,000 MNIST digits, unseen during training, of class 0 (blue) and otherwise (grey). The automaton produced by training in variable-depth mode induces a bimodal distribution of dynamical activities for each digit class.}
   \label{fig1}
\end{figure}

\begin{figure*}[] 
   \centering
  \includegraphics[width=\linewidth]{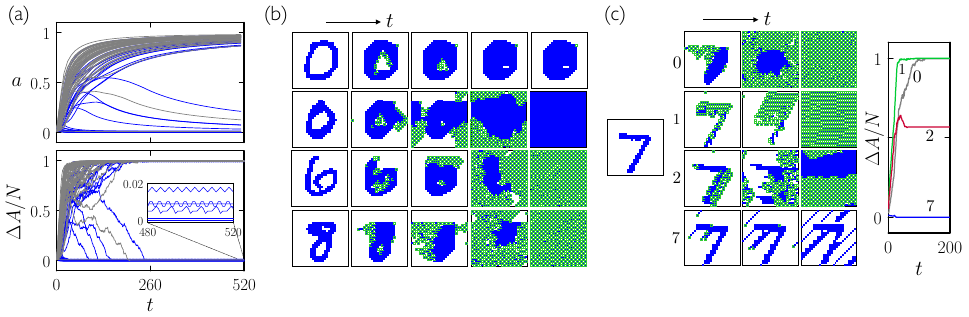} 
   \caption{(a) Time-integrated activity (top), \eqq{ay}, and its instantaneous counterpart (bottom), \eqq{ayinst}, as a function of time, for the automaton discovered by the variable-depth search of \f{fig1}. Shown are 100 of the 60,000 trajectories comprising the histogram of the bottom panel of \f{fig1}(b). Trajectories begun from digits of class 0 are blue. (b) Time-ordered snapshots generated by the automaton when presented with 4 different images. Cells with green boundaries have changed state from the preceding time step. (c) Time-ordered snapshots (left) and instantaneous activity as a function of time (right) for 4 automata. Each was trained in variable-depth mode to propagate an inactive trajectory when presented with a digit of the indicated class, and an active trajectory otherwise. Each is presented with the same digit, a 7 (shown left).}
   \label{fig2}
\end{figure*}

 The table ${\cal R}(s)$ thus has 512 entries, indicating the state $S_i(t+1)$ of cell $i$ at time $t+1$ given the state $s_i(t)$ of its environment at time $t$. Each entry in the table is 0 or 1, and so there are $2^{512} \sim 10^{154}$ possible rules\c{packard1985two}. We start with the identity rule table, for which $S_i(t+1)=S_i(t)$. One application of the table to each cell on the lattice constitutes one step of the automaton. The total number of steps $T$ we call the depth of the automaton.

To search for automata able to classify images we selected a subset (a minibatch) of $M=5000$ images from the MNIST training set, and did zero-temperature Metropolis Monte Carlo search~\footnote{Also known as random-mutation hill climbing\c{mitchell1993will,mitchell1998introduction}.} on the rule table. This procedure, shown schematically in \f{fig_schematic}(c), proceeds as follows~\footnote{This approach is similar to those that use genetic algorithms to search for automaton rules that perform particular computations\c{suzudo2004searching,mitchell1996evolving,oliveira2009some,chavoya2006using}, except that the present search scheme applies mutations and the Metropolis acceptance criterion to a single individual, rather than applying genetic operations to a population of individuals. For continuous variables and small mutations, the present procedure constitutes noisy clipped gradient descent on the loss surface\c{whitelam2021correspondence}.}. Starting from the identity rule table and an automaton of depth $T$, we calculate an order parameter $\phi$ (described shortly) for the minibatch. We then propose a change of the 512-entry rule table ${\cal R}$ in $k$ positions. We alternate, every Monte Carlo step, between the choices $k=1$ and $k$ a uniform random number on $[1,10]$; in \f{fig_schematic}(c), we show a change to a single entry of the rule table. Using this new rule table we calculate the new value of \eq{phi}, called $\phi'$. If $\phi' \geq \phi$ then the change to ${\cal R}$ is accepted: the proposed rule table becomes the current rule table, and $\phi$ is set equal to $\phi'$. Otherwise, the original rule table and value of $\phi$ are retained.

We trained automata in two modes: fixed-depth mode, with $T$ fixed to 10, and variable-depth mode. In the latter, we set $T=2$ initially, and propose a change $T \to T\pm1$ (with equal likelihood) every 10 Monte Carlo steps. When proposing a change of $T$, no change is made to the rule table. In variable-depth mode, both the automaton rule table ${\cal R}$ and its depth $T$ are trainable parameters.

The fixed-depth search scheme provides a comparison for the variable-depth search scheme. The results of the latter show the emergence of trajectory phase coexistence when the depth $T$ becomes sufficiently large. We did not anticipate the existence of the phase-coexistence behavior, and nor, for a given image class and given realization of the Monte Carlo search scheme, do we know in advance how large $T$ must be to allow this behavior. Knowing the existence of this behavior we can attempt to reproduce it by training at various large fixed depths, but the variable-depth scheme removes the need for such guesswork. It also emphasizes that the action of an automaton is a function of both its rule table and its depth.

To classify images we need a suitable order parameter $\phi$. One possibility is to consider the state of a given cell\c{gers1997codi} or to count types of local environment, but these break translational invariance and are biased toward certain image types, respectively. An alternative, borrowed from studies of glasses, is {\em activity}, the total number of changes of state of all cells in the course of a simulation. Activity is a simple measure that does not affect translational invariance and is not biased toward any particular spatial pattern. If $S_i(t)$ is the state of cell $i$ at time $t$, then we define activity
\beq 
\label{act}
A(T)\equiv \sum_{t=0}^{T-1} \sum_{i=1}^N \left(1-\delta_{S_i(t),S_i(t+1)} \right)
\eeq
as the total number of changes of state of all cells under the action of the automaton. Here $\delta_{x,y}$ is the Kronecker delta, equal to 1 if $x=y$ and 0 otherwise. We further define $A_j(T)$ as the value of \eq{act} upon starting from image $j$, i.e. upon setting $S_i(0) = B_{i,j}$ for all $i$, where $B_{i,j}$ is the binary pixel pattern of image $j$ (shown for one image in \f{fig_schematic}(a)).  Let 
\beq
\label{ay}
a_j(T) \equiv \frac{A_j(T)}{NT}\in [0,1]
\eeq
be the scaled counterpart of $A_j(T)$. We then define the order parameter
\beq
\label{phi}
\phi \equiv \frac{1}{M_{{\bar{\alpha}}}} \sum_{C(j) \neq \alpha} a_j(T) -\frac{1}{M_\alpha} \sum_{C(j) = \alpha} a_j(T),
\eeq
where the first sum runs over all $M_{{\bar{\alpha}}} \equiv M-M_\alpha$ images not of class $\alpha$ in the minibatch, and the second sum runs all $M_\alpha$ images of class $\alpha$ ($C(j)$ returns the class of image $j$). Note that $\phi$ is a property of the minibatch, not of individual automaton trajectories, and has range $[-1,1]$ in principle and $[0,1]$ in practice, given the initial conditions and acceptance criterion of the search scheme. The instruction to maximize \eq{phi} is the instruction to find an automaton whose dynamics is inactive if initiated from an image of class $\alpha$, and as active as possible otherwise.

We note that the model and simulation algorithm are translationally invariant, but not rotationally invariant (by design, because the class of a number is not invariant to rotation). Rotational invariance (and other symmetries) could be imposed during the search scheme, by limiting the search to automaton rules that respect those symmetries. Here we carry out a general search, with no restrictions.

\section{Classification by dynamical phase coexistence} 
\label{pc}

In \f{fig1} we show results of two instances of the Monte Carlo search scheme for image class $\alpha=0$ ; these are typical of results for other image classes. The fixed-depth automaton learns to increase the value of $\phi$, shown in panel (a), with Monte Carlo step number $n$. The associated histograms of activity for 60,000 binarized MNIST digits not seen during training are shown in panel (b), top. Symbols indicate nonzero values in windows of width $1/50$, while lines are a guide to the eye. The sum of symbol values for each histogram is unity. The distribution of activity for each digit class is unimodal, and class 0 (blue lines and symbols) is on average less active than the others (grey lines and symbols; image class $\alpha \neq 0$ is denoted by grey polygons with $\alpha+3$ sides). There is considerable overlap between classes.

Automata trained by the variable-depth search scheme behave differently. About one-third of these adopted a depth of 1, but in a majority of cases we observed behavior of the kind shown in the figure. The order parameter soon exceeds that of the fixed-depth scheme, and the depth of the automaton grows steadily, exceeding 500 timesteps within the period allotted to training (\f{fig1}(a)). The activity distributions also differ from those of the fixed-depth scheme, being bimodal for each class (\f{fig1}(b)). Zeros are considerably more likely to give rise to inactive trajectories than active ones, while the opposite is true of the other classes, although exceptions are numerous (we will return to the effectiveness of these rules as classifiers). The inset shows histograms generated using the same automaton run for 3000 timesteps, revealing that long trajectories begun from the MNIST digits display either high-activity or low-activity dynamics.

In \f{fig2}(a) we show 100 trajectories of the automaton discovered by variable-depth search when presented with 100 images. 50 images are of class 0 (blue lines) and 50 are of other classes (grey lines). The top panel shows the time-integrated scaled activity of each trajectory, \eqq{ay}, as a function of the number of automaton timesteps. Two distinct populations of trajectories are apparent, with the majority of zeros belonging to the low-activity population. The bottom panel shows, for the same trajectories, the instantaneous activity 
\beq
\label{ayinst}
\Delta A_j(t) \equiv A_j(t)-A_j(t-1)
\eeq
accrued at timestep $t$. This format shows that trajectories tend, after about 250 timesteps, to persistently inactive or active solutions. In the inset to the bottom figure we show some of the low-activity trajectories on a larger scale. Some of the solutions adopted are absorbing states, having zero activity, some have persistent small values of activity, and some solutions are periodic, moving between configurations that generate slightly different values of activity. Trajectories thus show a range of behavior, and do not all correspond to the same values of activity, but nonetheless divide naturally into two populations or ``phases'' whose order-parameter values are tightly distributed and well separated from each other. The time-integrated order parameter shown in the top panel of \f{fig2}(a) will, in the limit of infinite time, adopt the value consistent with the limiting form of the instantaneous order parameter $\Delta A$ shown in the bottom panel. 

The bimodal dynamical behavior shown in \f{fig2}(a) is similar to that seen in systems with explicit feedback or memory\c{klymko2017similarity,whitelam2021varied}, and resembles an emergent version of a spiking neuron\c{paugam2012computing,gerstner2002spiking}. Spiking neurons are time-dependent models of biological neurons, often modeled by differential equations, whose outputs can vary sharply with time in response to a stimulus. Here the spiking results from the many-body dynamics of the automaton: it performs a time-dependent computation from an initial condition specified by an image, and (with some error) becomes asymptotically quiescent if it detects a specified image type and spikes otherwise. Cellular automata resemble deep neural networks in the sense that they enact a nonlinear transformation by propagating a dynamics or a computer program as a function of time or depth\c{schmidhuber1997discovering,schmidhuber2015deep,goodfellow2016deep}. However, neural-net weights usually vary with depth, making a single automaton rule closer in sense to a recurrent neural network, or to a single activation function that performs a time-dependent computation.

In \f{fig2}(b) we show time-ordered snapshots of the automaton dynamics starting from 4 different images. Blue and white pixels denote the two automaton cell states. Cells with a green boundary are active, meaning that their state in the previous timestep was different, while cells not labeled in this way are inactive, having the same state as in the previous timestep. The top panel shows the automaton identifying a zero by propagating a low-activity trajectory. It performs what is in effect a local computation, causing the image to become increasingly compact until it adopts a state in which only one cell on the periphery of the image remains persistently active, changing state every timestep. The second row shows the automaton identifying a different zero by propagating an asymptotically inactive trajectory, but in this case the computation is carried out across the whole lattice, and involves a long transient. One part of the image gives rise to an active pattern and another to an inactive pattern, and these coexist for some time before the inactive pattern consumes the lattice. The bottom two rows show the automaton identifying images that are not zeros, by propagating active trajectories. In both cases parts of the images serve as nucleation sites for active patterns, which, after some time, consume the inactive portions of the lattice. A similar dynamics is seen when the automaton misclassifies an image: in those cases, the competition between active and inactive portions of the lattice is won by the ``wrong'' phase.

In \f{fig2}(c) we show time-ordered snapshots of four different automata~\footnote{Times of snapshots for each row are as follows. 0: $10,38,200$; 1: $3,9,200$; 2: $3,13,200$; 7: $5,10,200$.}. Each automaton was trained, in variable-depth mode, to generate an inactive trajectory when presented with an image of the indicated class, and an active trajectory otherwise. In each case the depths identified by the training process were several hundred timesteps. All are presented with the same digit, a 7. The plot at right shows the activity accrued at each timestep, \eqq{ayinst}. The 0- and 1-automata generate active trajectories, while the 7-automaton generates an inactive one. The 2-automaton illustrates a minority behavior seen in some of the automata, which display a bimodal activity distribution peaked at high and low values but with some small fraction of trajectories having intermediate values. This example falls in the latter category. The automaton produces an ambiguous answer in which, at long times, high- and low-activity patterns coexist in a spatial sense, similar to behavior seen in automata such as models of traffic\c{jiang2002cellular,wolf1999cellular,neto2011phase,d2005coexisting} or the Ising model\c{kinzel1985phase,binney1992theory}. 

In \s{examples} we provide the rule tables for three automata trained to recognize different digit classes. 

\section{Reservoir computing with automata}
\label{acc}

A single automaton has a modest ability to classify images: for instance, the automaton obtained by variable-depth training in \f{fig1} classifies zeros with about $2/3$ accuracy. In addition, distinct automata trained using the same procedure can identify specific digits differently, as shown in \f{fig3}. There we show the action of 50 trained automata, 5 per class, on 3 MNIST digits. A blue square indicates that the automaton propagates an inactive trajectory (defined as $a_j(T)<1/4$) when presented with the digit $j$ shown at left, and so recognizes it as a member of the class it is trained to identify. A grey square indicates the opposite. In these examples all the automata that should recognize the digit as a member of their own classes do so, but false positives are common. For example, the 4-automata all identify the 1 as a 4, and the 0 is incorrectly recognized as a member of their own class by automata of 6 other classes.

\begin{figure}[] 
   \centering
  \includegraphics[width=0.8\linewidth]{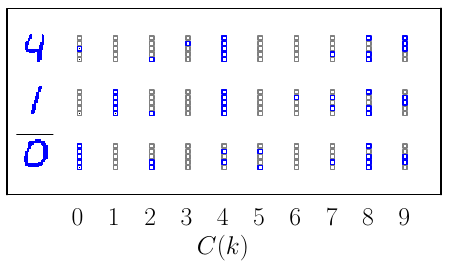} 
   \caption{Action on 3 images (left) of 50 automata $k$ (squares) trained to recognize image class $C(k)$. The presence of a blue square indicates that the automaton recognizes the image by propagating a trajectory belonging to its low-activity dynamical phase.}
   \label{fig3}
\end{figure}

Acting in concert, automata are considerably more accurate. To illustrate this fact we adopt the approach of reservoir computing, in which a nonlinear system or reservoir -- which can be a neural network, a cellular automaton, or even a physical system such as a bucket of water -- enacts a transformation on a set of data, before a final layer, usually a linear model, is trained in order to do computation\c{yilmaz2015machine,nichele2017reservoir,moran2018reservoir,tanaka2019recent}. In reservoir computing the reservoir is not usually trained, obviating the need for doing gradient descent on a complex nonlinear system. Here we use as a reservoir a set of cellular automata, each trained by Monte Carlo search, and pass their output to a final layer that is subsequently trained.

We proceed as follows. We construct $N_{\rm f}$ cellular automata using the variable-depth search algorithm described in the previous section. We considered the original order parameter $\phi$, \eqq{phi}, and 5 others, described below. Automata were trained on 5000 digits of the MNIST training set, and training was terminated after 5 CPU hours. Any automaton whose depth was unity after 500 Monte Carlo steps was reset (to the identity rule table and depth 2). The automata produced in this way are heterogenous, some exhibiting the phase coexistence-like behavior discussed previously, and some remaining relatively shallow and exhibiting unimodal activity distributions. 

One trained, we pass all 70,000 MNIST digits through all $N_{\rm f}$ automata. For each digit we consider the $N_{\rm f}$ values of activity that result to comprise an $N_{\rm f}$-dimensional feature vector. We train a final layer, using these feature vectors, on the 60,000 digits of the MNIST training set. With this trained layer we calculate $A$, its classification accuracy on the 10,000 digits of the MNIST test set.

We considered two versions of the final layer. The first is a logistic (generalized linear) model, conventional in reservoir computing. This has 
\beq
\label{nlc}
N_{\rm lc}(N_{\rm f}) = (N_{\rm f} +1)C
\eeq 
real-valued parameters, where $C=10$ is the number of digit classes. For a given digit its output is a $C$-dimensional vector whose entries correspond to its prediction for the class of the digit. We trained this model using multinomial regression. The second was a nonlinear model, a single-layer neural network with 128 hidden neurons, each with tanh nonlinearities. This model has 
\beq
\label{nnet}
N_{\rm net}(N_f)= (128+C) N_f+128+C
\eeq 
real-valued parameters. For a given digit its output is also a $C$-dimensional vector whose entries correspond to its prediction for the class of the digit. We trained this model using the Adam optimizer\c{kingma2014adam} in PyTorch\c{paszke2019pytorch} (using minibatches of size 1000 we trained for 200 epochs with a learning rate of $10^{-3}$).

We carried out this exercise using six different reservoirs, each trained by Monte Carlo search to maximize one of 6 order parameters, in order to understand how important is the choice of order parameter to the ability of the final layer to do classification. To introduce these order parameters we define the average
\beq
\av{\cdot}_\alpha \equiv \frac{1}{M_\alpha} \sum_{C(j) = \alpha} \left( \cdot \right),
\eeq
the sum running over all $M_\alpha$ images of class $\alpha$ in the minibatch, and 
\beq
\av{\cdot}_{\bar{\alpha}} \equiv \frac{1}{M_{\bar{\alpha}}} \sum_{C(j) \neq \alpha} \left( \cdot \right),
\eeq
the sum running over all $M_{{\bar{\alpha}}} \equiv M-M_\alpha$ images not of class $\alpha$ in the minibatch. The first order parameter considered is the original, \eqq{phi}. The second is its sign-reversed counterpart,
\beq
\label{phi2}
\phi_2 = \av{a_j(T)}_\alpha - \av{a_j(T)}_{\bar{\alpha}},
\eeq
the instruction to maximize which is the instruction to maximize the activity of image class $\alpha$ and minimize the activity of all other classes. 

The third is 
\beq
\label{phi3}
\phi_3 = |\av{a_j(T)}_\alpha -  \av{a_j(T)}_{\bar{\alpha}}|,
\eeq
which encourages separation of the mean activity of class $\alpha$ from the mean of the other $9$ image classes, but does not specify whether the activity of class $\alpha$ should be greater than or less than those of the other classes. 

The fourth is the preceding order parameter with an additional constraint on the standard deviation $\sigma_\alpha \equiv \sqrt{\av{a_j(T)^2}_\alpha-\av{a_j(T)}_\alpha^2}$ of the distributions of the chosen class and its complement:
\beq
\label{phi4}
\phi_4=\phi_3 - \frac{1}{2}( \sigma_\alpha +  \sigma_{\bar \alpha}).
\eeq

For each of the order parameters $\phi$ to $\phi_4$ we constructed reservoirs in which automata instructed to identify each class were equally numerous.

The fifth order parameter is a version of $\phi_4$ that rewards separation of all $C=10$ classes and the minimization of the variance of their distributions:
\bea
\label{phi5}
\phi_5&=&\frac{1}{P} \sum_{\alpha \neq \beta} |\av{a_j(T)}_\alpha -  \av{a_j(T)}_{\beta}|-\frac{1}{C} \sum_{\alpha=1}^C\sigma_\alpha,
\eea
the first sum running over all $P=C(C-1)/2=45$ pairs of classes and the second running over all $C=10$ classes.

The final order parameter is the analog of $\phi_5$ for two class types, $\alpha$ and $\beta$,
\bea
\label{phi6}
\phi_6&=&\ |\av{a_j(T)}_\alpha -  \av{a_j(T)}_{\beta}|-\frac{1}{2}( \sigma_\alpha +  \sigma_{\beta}).
\eea
When constructing reservoirs using $\phi_6$ as an order parameter we assign each automaton at random to one of the 45 class-pairs $(\alpha,\beta)$. Each used only digits of those two classes from the 5000-digit minibatch, approximately 1000 digits in all cases.

\begin{table}[]
\caption{Reservoir computing with trained automata. MNIST is binarized and passed through $N_{\rm f}$ automata, each trained using the order parameter $\phi_i$. The resulting activities constitute a set of $N_{\rm f}$-dimensional feature vectors on which we train a logistic model (log) or a fully-connected neural network (net). $A$ is the resulting MNIST test-set accuracy percentage, with parameter memory cost in kilobytes in brackets.}
\centering
\begin{tabular}{cccc}
\hline
scheme & $N_{\rm f}$/features & $A_{\rm log}$ (mem) & $A_{\rm net}$ (mem)\\ 
\hline
MNIST & 784 & 92.6 (31) & 98.0 (433) \\
binarized MNIST & 784 & 92.1 (31) & 97.7 (433) \\
\hline
$\phi$ & 500 & 95.3 (52) & 96.3 (309) \\
$\phi_2$ & 500 & 95.4 (52)& 96.2 (309) \\
$\phi_3$ & 500 & 95.3 (52)& 96.3 (309) \\
$\phi_4$ & 500 & 96.8 (52) & 97.2 (309) \\
$$ & 870 & 97.3 (91) & 98.1 (537) \\
$\phi_5$ & 500 & 94.6 (52) & 95.3 (309)\\
$\phi_6$ & 80 & 91.9 (8)  &95.2 (50) \\
$$ & 255 & 95.7 (27) &96.9 (158) \\
$$ & 500 & 96.7 (52) &97.3 (309)\\
all & 3370 &97.7 (350) & 98.1 (2076)\\
\hline
untrained & 500 & 88.0 (52) & 93.0 (309) \\
\hline
\end{tabular}
\label{tab1}
\end{table}

Results are shown in Table~\ref{tab1}. For reference, the first two rows show classification accuracy $A$ of the two versions of the final layer using the original MNIST feature vectors and MNIST binarized using the threshold (1/4 of the maximum pixel intensity) used by the automata. Some small loss of accuracy is incurred by binarizing MNIST.

We shall focus first on results using the logistic model as a final layer, shown in the third column of the table. The order parameters $\phi$ to $\phi_3$, which encourage automata to make the dynamics of one digit class more or less active than the others, perform similarly. Reservoirs of 500 automata each provide about 3\% improvement in logistic-model classification accuracy over the original MNIST data, indicating that the nonlinear transformations enacted by the automata are beneficial, separating classes in phase space in a way that a logistic model finds easier to process. The three order parameters provide essentially the same numerical benefit, emphasizing that activity is an order parameter agnostic to digit type: it does not matter if automata are instructed to make certain classes more or less active than others.

Moving from the order parameter $\phi_3$ to $\phi_4$, we see that some additional numerical improvement is afforded by the extra terms in $\phi_4$ that encourage automata to produce narrow distributions of activity. Moreover, more reservoir automata generally equates to better performance, and it is possible to construct reservoirs possessing more members than there are features in MNIST: with 870 features the logistic-model accuracy exceeds by about 5\% the corresponding number using the original data set.

The order parameter $\phi_5$, which encourages each automaton to separate activity distributions for all classes, performs less well than the other order parameters. $\phi_5$ is the most difficult order parameter for an individual automaton to maximize, and this difficulty results in a reservoir slightly less expressive than those made using ``easier'' order parameters.

The order parameter $\phi_6$, which encourages separation of two classes and ignores the rest, performs similarly to $\phi_4$. We show also that classification accuracy increases with increasing reservoir size. Indeed, the accuracy of the logistic model increases further if presented with all of the previous reservoirs combined, shown in the penultimate row of the table.

As a point of comparison with the trained reservoirs, we show in the final row of the table the performance of a reservoir of 500 automata whose rules were chosen randomly, as is conventional in reservoir computing. For each, we drew a random number $\lambda$ uniformly on $(0,1]$, and populated each entry of its rule table with a 1 with probability $\lambda$ and a 0 otherwise ($\lambda$ is then the mean fraction of 1s in the rule table\c{langton1990computation}). We trained the depth of each automaton by starting at $T=1$ and increasing $T$ by 1 until the order parameter $\phi_5$ began to decrease. This reservoir of randomly-chosen rules performs less well with the linear classifier than does the original data set, indicating that information has been lost by the reservoir and highlighting the utility of training the rules of each automaton.

The performance of the nonlinear final layer, shown in the fourth column of Table~\ref{tab1}, is different to that of the logistic model. With the neural network, the best reservoirs and the original data set yield similar results. Comparison of the performance of the two versions of the final layer indicates that the nonlinear transformations enacted by the reservoir are unnecessary for the neural network, which is expressive enough that it can extract similar information using the original data set. 

We turn finally to the memory consumption of each method. We represent real-valued parameters with 32-bit precision, and binary variables, which specify the automata rule tables, cost 1 bit. Thus for $N_{\rm f}$ features the linear classifier costs $32\,  {\rm bits} \times N_{\rm lc}(N_{\rm f})$, given by \eqq{nlc}, and the net costs $32\,  {\rm bits} \times N_{\rm net}(N_{\rm f})$, given by \eqq{nnet}, while each automaton costs $1\,  {\rm bit} \times 512$. The resulting memory cost of the methods (to the nearest kilobyte; 8 bits = 1 byte) are shown in brackets in the table. For example, using binarized MNIST, the net achieves an accuracy of 97.7\% at a memory cost of 433 kb; all the reservoirs connected to a linear classifier achieve 97.7\%, with a memory cost of 350 kb. 870 automata trained using the order parameter $\phi_4$ achieve the slightly lower accuracy of 97.3\%, but at a reduced memory cost of 91 kb. This is equivalent to about 22,750 real-valued parameters, and we can assess the accuracy of similarly-sized fully-connected networks acting directly on MNIST or binarized MNIST. 22,750 parameters equates to a single-layer fully-connected net of about 28 hidden nodes, which scores 96.4\% (MNIST) or 95.7\% (binarized MNIST); or to a two-layer net with about 27 nodes per layer (96.2\% or 95.9\%); or to a three-layer net with about 26 nodes per layer (96.1\% or 95.6\%); or to a four-layer net with about 25 nodes per layer (96.2\% or 95.1\%); or to a ten-layer net with about 22 nodes per layer (94.6\% or 93.5\%). Thus a linear classifier acting on a set of automata can achieve comparable accuracy to fully-connected neural networks acting directly on MNIST, with a smaller memory cost. Both are less accurate than certain other types of architecture, such as convolutional neural networks\c{lecun1995learning,mnist}.

\section{Conclusions}
\label{conc}

We have introduced dynamical activity, an order parameter with no bias toward any particular environment or spatial position, as a metric for image classification by cellular automata. Zero-temperature Metropolis Monte Carlo search of rules and depths for general 2D automata has identified algorithms that classify MNIST images by enacting a form of dynamical phase coexistence, propagating trajectories belonging to high- or low-activity dynamical phases. Individual automata achieve this classification with modest accuracy. Collections or reservoirs of such automata can be used by a linear model to classify images with comparable accuracy (and smaller memory cost) than fully-connected neural networks that act on the pixels of the original image.

Conceptually, the automata described here evoke concepts or display features seen in other studies, but in detail are distinct from automata described previously. For example, several authors have discussed the computational potential of automaton rules close to phase transitions in rule space\c{wootters1990there,li1990transition,mitchell1994dynamics,langton1990computation}. We can view the transition between the behaviors shown in the top and bottom panels of \f{fig1}(b) as a phase transition in rule space, analogous to the transition shown by the Ising model as its temperature is reduced below the critical value. However, the defining character of the automata described here is that they do computation by enacting a form of dynamical phase coexistence, and so the computation is enabled by a phase transition taking place in trajectory space, not rule space.

In addition, automata of this nature display elements of the 4 Wolfram classes of cellular automata\c{wolfram1984universality}, but are not completely described by these rules. For example, \f{fig2} shows that a single automaton, starting from different initial conditions, can generate a trajectory similar to that of a Class 1 automaton (ending in a homogenous state, i.e. a limit point), or a Class 2 automaton (generating structures periodic in time, i.e. limit cycles), and can generate complex patterns with long transients. In these respects it resembles a Class 4 automaton, which can generate both repetitive and complex patterns, but the defining characteristic of the automaton is that its dynamical ensemble contains trajectories of two distinct types, defined by their activity, a concept separate from those considered in the Wolfram scheme. Indeed, the dynamical phase coexistence seen in these automata resembles that seen in some stochastic systems\c{kinzel1985phase,binney1992theory,klymko2017similarity,hedges2009dynamic}, a connection of the kind anticipated in Question 10 (``What is the correspondence between cellular automata and
stochastic systems?'') of~\cc{wolfram1985twenty}.

{\em Acknowledgments.---} This work was performed as part of a user project at the Molecular Foundry, Lawrence Berkeley National Laboratory, supported by the Office of Science, Office of Basic Energy Sciences, of the U.S. Department of Energy under Contract No. DE-AC02--05CH11231. I.T. acknowledges NSERC.
 

\begin{thebibliography}{52}%
\makeatletter
\providecommand \@ifxundefined [1]{%
 \@ifx{#1\undefined}
}%
\providecommand \@ifnum [1]{%
 \ifnum #1\expandafter \@firstoftwo
 \else \expandafter \@secondoftwo
 \fi
}%
\providecommand \@ifx [1]{%
 \ifx #1\expandafter \@firstoftwo
 \else \expandafter \@secondoftwo
 \fi
}%
\providecommand \natexlab [1]{#1}%
\providecommand \enquote  [1]{``#1''}%
\providecommand \bibnamefont  [1]{#1}%
\providecommand \bibfnamefont [1]{#1}%
\providecommand \citenamefont [1]{#1}%
\providecommand \href@noop [0]{\@secondoftwo}%
\providecommand \href [0]{\begingroup \@sanitize@url \@href}%
\providecommand \@href[1]{\@@startlink{#1}\@@href}%
\providecommand \@@href[1]{\endgroup#1\@@endlink}%
\providecommand \@sanitize@url [0]{\catcode `\\12\catcode `\$12\catcode
  `\&12\catcode `\#12\catcode `\^12\catcode `\_12\catcode `\%12\relax}%
\providecommand \@@startlink[1]{}%
\providecommand \@@endlink[0]{}%
\providecommand \url  [0]{\begingroup\@sanitize@url \@url }%
\providecommand \@url [1]{\endgroup\@href {#1}{\urlprefix }}%
\providecommand \urlprefix  [0]{URL }%
\providecommand \Eprint [0]{\href }%
\providecommand \doibase [0]{http://dx.doi.org/}%
\providecommand \selectlanguage [0]{\@gobble}%
\providecommand \bibinfo  [0]{\@secondoftwo}%
\providecommand \bibfield  [0]{\@secondoftwo}%
\providecommand \translation [1]{[#1]}%
\providecommand \BibitemOpen [0]{}%
\providecommand \bibitemStop [0]{}%
\providecommand \bibitemNoStop [0]{.\EOS\space}%
\providecommand \EOS [0]{\spacefactor3000\relax}%
\providecommand \BibitemShut  [1]{\csname bibitem#1\endcsname}%
\let\auto@bib@innerbib\@empty
\bibitem [{\citenamefont {Von~Neumann}(1968)}]{von2017general}%
  \BibitemOpen
  \bibfield  {author} {\bibinfo {author} {\bibfnamefont {John}\ \bibnamefont
  {Von~Neumann}},\ }\bibfield  {title} {\enquote {\bibinfo {title} {The general
  and logical theory of automata},}\ }in\ \href@noop {} {\emph {\bibinfo
  {booktitle} {Systems Research for Behavioral Sciencesystems Research}}}\
  (\bibinfo  {publisher} {Routledge},\ \bibinfo {year} {1968})\BibitemShut
  {NoStop}%
\bibitem [{\citenamefont {Conway}\ \emph {et~al.}(1970)\citenamefont {Conway}
  \emph {et~al.}}]{conway1970game}%
  \BibitemOpen
  \bibfield  {author} {\bibinfo {author} {\bibfnamefont {John}\ \bibnamefont
  {Conway}} \emph {et~al.},\ }\bibfield  {title} {\enquote {\bibinfo {title}
  {The game of life},}\ }\href@noop {} {\bibfield  {journal} {\bibinfo
  {journal} {Scientific American}\ }\textbf {\bibinfo {volume} {223}},\
  \bibinfo {pages} {4} (\bibinfo {year} {1970})}\BibitemShut {NoStop}%
\bibitem [{\citenamefont {Wolfram}(1984)}]{wolfram1984universality}%
  \BibitemOpen
  \bibfield  {author} {\bibinfo {author} {\bibfnamefont {Stephen}\ \bibnamefont
  {Wolfram}},\ }\bibfield  {title} {\enquote {\bibinfo {title} {Universality
  and complexity in cellular automata},}\ }\href@noop {} {\bibfield  {journal}
  {\bibinfo  {journal} {Physica D: Nonlinear Phenomena}\ }\textbf {\bibinfo
  {volume} {10}},\ \bibinfo {pages} {1--35} (\bibinfo {year}
  {1984})}\BibitemShut {NoStop}%
\bibitem [{\citenamefont {Wolfram}(1983)}]{wolfram1983statistical}%
  \BibitemOpen
  \bibfield  {author} {\bibinfo {author} {\bibfnamefont {Stephen}\ \bibnamefont
  {Wolfram}},\ }\bibfield  {title} {\enquote {\bibinfo {title} {Statistical
  mechanics of cellular automata},}\ }\href@noop {} {\bibfield  {journal}
  {\bibinfo  {journal} {Reviews of modern physics}\ }\textbf {\bibinfo {volume}
  {55}},\ \bibinfo {pages} {601} (\bibinfo {year} {1983})}\BibitemShut
  {NoStop}%
\bibitem [{\citenamefont {Chopard}\ and\ \citenamefont
  {Droz}(1998)}]{chopard1998cellular}%
  \BibitemOpen
  \bibfield  {author} {\bibinfo {author} {\bibfnamefont {B}~\bibnamefont
  {Chopard}}\ and\ \bibinfo {author} {\bibfnamefont {M}~\bibnamefont {Droz}},\
  }\href@noop {} {\emph {\bibinfo {title} {Cellular automata}}},\ Vol.~\bibinfo
  {volume} {1}\ (\bibinfo  {publisher} {Springer},\ \bibinfo {year}
  {1998})\BibitemShut {NoStop}%
\bibitem [{\citenamefont {Sant{\'e}}\ \emph {et~al.}(2010)\citenamefont
  {Sant{\'e}}, \citenamefont {Garc{\'\i}a}, \citenamefont {Miranda},\ and\
  \citenamefont {Crecente}}]{sante2010cellular}%
  \BibitemOpen
  \bibfield  {author} {\bibinfo {author} {\bibfnamefont {In{\'e}s}\
  \bibnamefont {Sant{\'e}}}, \bibinfo {author} {\bibfnamefont {Andr{\'e}s~M}\
  \bibnamefont {Garc{\'\i}a}}, \bibinfo {author} {\bibfnamefont {David}\
  \bibnamefont {Miranda}}, \ and\ \bibinfo {author} {\bibfnamefont {Rafael}\
  \bibnamefont {Crecente}},\ }\bibfield  {title} {\enquote {\bibinfo {title}
  {Cellular automata models for the simulation of real-world urban processes: A
  review and analysis},}\ }\href@noop {} {\bibfield  {journal} {\bibinfo
  {journal} {Landscape and urban planning}\ }\textbf {\bibinfo {volume} {96}},\
  \bibinfo {pages} {108--122} (\bibinfo {year} {2010})}\BibitemShut {NoStop}%
\bibitem [{\citenamefont {Ganguly}\ \emph {et~al.}(2002)\citenamefont
  {Ganguly}, \citenamefont {Maji}, \citenamefont {Dhar}, \citenamefont
  {Sikdar},\ and\ \citenamefont {Chaudhuri}}]{ganguly2002evolving}%
  \BibitemOpen
  \bibfield  {author} {\bibinfo {author} {\bibfnamefont {Niloy}\ \bibnamefont
  {Ganguly}}, \bibinfo {author} {\bibfnamefont {Pradipta}\ \bibnamefont
  {Maji}}, \bibinfo {author} {\bibfnamefont {Sandip}\ \bibnamefont {Dhar}},
  \bibinfo {author} {\bibfnamefont {Biplab~K}\ \bibnamefont {Sikdar}}, \ and\
  \bibinfo {author} {\bibfnamefont {P~Pal}\ \bibnamefont {Chaudhuri}},\
  }\bibfield  {title} {\enquote {\bibinfo {title} {Evolving cellular automata
  as pattern classifier},}\ }in\ \href@noop {} {\emph {\bibinfo {booktitle}
  {International Conference on Cellular Automata}}}\ (\bibinfo {organization}
  {Springer},\ \bibinfo {year} {2002})\ pp.\ \bibinfo {pages}
  {56--68}\BibitemShut {NoStop}%
\bibitem [{\citenamefont {Maji}\ \emph {et~al.}(2003)\citenamefont {Maji},
  \citenamefont {Shaw}, \citenamefont {Ganguly}, \citenamefont {Sikdar},\ and\
  \citenamefont {Chaudhuri}}]{maji2003theory}%
  \BibitemOpen
  \bibfield  {author} {\bibinfo {author} {\bibfnamefont {Pradipta}\
  \bibnamefont {Maji}}, \bibinfo {author} {\bibfnamefont {Chandrama}\
  \bibnamefont {Shaw}}, \bibinfo {author} {\bibfnamefont {Niloy}\ \bibnamefont
  {Ganguly}}, \bibinfo {author} {\bibfnamefont {Biplab~K}\ \bibnamefont
  {Sikdar}}, \ and\ \bibinfo {author} {\bibfnamefont {P~Pal}\ \bibnamefont
  {Chaudhuri}},\ }\bibfield  {title} {\enquote {\bibinfo {title} {Theory and
  application of cellular automata for pattern classification},}\ }\href@noop
  {} {\bibfield  {journal} {\bibinfo  {journal} {Fundamenta Informaticae}\
  }\textbf {\bibinfo {volume} {58}},\ \bibinfo {pages} {321--354} (\bibinfo
  {year} {2003})}\BibitemShut {NoStop}%
\bibitem [{\citenamefont {Yilmaz}(2015)}]{yilmaz2015machine}%
  \BibitemOpen
  \bibfield  {author} {\bibinfo {author} {\bibfnamefont {Ozgur}\ \bibnamefont
  {Yilmaz}},\ }\bibfield  {title} {\enquote {\bibinfo {title} {Machine learning
  using cellular automata based feature expansion and reservoir computing.}}\
  }\href@noop {} {\bibfield  {journal} {\bibinfo  {journal} {Journal of
  Cellular Automata}\ }\textbf {\bibinfo {volume} {10}} (\bibinfo {year}
  {2015})}\BibitemShut {NoStop}%
\bibitem [{\citenamefont {Nichele}\ and\ \citenamefont
  {Gundersen}(2017)}]{nichele2017reservoir}%
  \BibitemOpen
  \bibfield  {author} {\bibinfo {author} {\bibfnamefont {Stefano}\ \bibnamefont
  {Nichele}}\ and\ \bibinfo {author} {\bibfnamefont {Magnus~S}\ \bibnamefont
  {Gundersen}},\ }\bibfield  {title} {\enquote {\bibinfo {title} {Reservoir
  computing using non-uniform binary cellular automata},}\ }\href@noop {}
  {\bibfield  {journal} {\bibinfo  {journal} {arXiv preprint arXiv:1702.03812}\
  } (\bibinfo {year} {2017})}\BibitemShut {NoStop}%
\bibitem [{\citenamefont {Mor{\'a}n}\ \emph {et~al.}(2018)\citenamefont
  {Mor{\'a}n}, \citenamefont {Frasser},\ and\ \citenamefont
  {Rossell{\'o}}}]{moran2018reservoir}%
  \BibitemOpen
  \bibfield  {author} {\bibinfo {author} {\bibfnamefont {Alejandro}\
  \bibnamefont {Mor{\'a}n}}, \bibinfo {author} {\bibfnamefont {Christiam~F}\
  \bibnamefont {Frasser}}, \ and\ \bibinfo {author} {\bibfnamefont {Josep~L}\
  \bibnamefont {Rossell{\'o}}},\ }\bibfield  {title} {\enquote {\bibinfo
  {title} {Reservoir computing hardware with cellular automata},}\ }\href@noop
  {} {\bibfield  {journal} {\bibinfo  {journal} {arXiv preprint
  arXiv:1806.04932}\ } (\bibinfo {year} {2018})}\BibitemShut {NoStop}%
\bibitem [{\citenamefont {Randazzo}\ \emph {et~al.}(2020)\citenamefont
  {Randazzo}, \citenamefont {Mordvintsev}, \citenamefont {Niklasson},
  \citenamefont {Levin},\ and\ \citenamefont {Greydanus}}]{randazzo2020self}%
  \BibitemOpen
  \bibfield  {author} {\bibinfo {author} {\bibfnamefont {Ettore}\ \bibnamefont
  {Randazzo}}, \bibinfo {author} {\bibfnamefont {Alexander}\ \bibnamefont
  {Mordvintsev}}, \bibinfo {author} {\bibfnamefont {Eyvind}\ \bibnamefont
  {Niklasson}}, \bibinfo {author} {\bibfnamefont {Michael}\ \bibnamefont
  {Levin}}, \ and\ \bibinfo {author} {\bibfnamefont {Sam}\ \bibnamefont
  {Greydanus}},\ }\bibfield  {title} {\enquote {\bibinfo {title}
  {Self-classifying mnist digits},}\ }\href@noop {} {\bibfield  {journal}
  {\bibinfo  {journal} {Distill}\ }\textbf {\bibinfo {volume} {5}},\ \bibinfo
  {pages} {e00027--002} (\bibinfo {year} {2020})}\BibitemShut {NoStop}%
\bibitem [{\citenamefont {Mor{\'a}n}\ \emph {et~al.}(2019)\citenamefont
  {Mor{\'a}n}, \citenamefont {Frasser}, \citenamefont {Roca},\ and\
  \citenamefont {Rossell{\'o}}}]{moran2019energy}%
  \BibitemOpen
  \bibfield  {author} {\bibinfo {author} {\bibfnamefont {Alejandro}\
  \bibnamefont {Mor{\'a}n}}, \bibinfo {author} {\bibfnamefont {Christiam~F}\
  \bibnamefont {Frasser}}, \bibinfo {author} {\bibfnamefont {Miquel}\
  \bibnamefont {Roca}}, \ and\ \bibinfo {author} {\bibfnamefont {Josep~L}\
  \bibnamefont {Rossell{\'o}}},\ }\bibfield  {title} {\enquote {\bibinfo
  {title} {Energy-efficient pattern recognition hardware with elementary
  cellular automata},}\ }\href@noop {} {\bibfield  {journal} {\bibinfo
  {journal} {IEEE Transactions on Computers}\ }\textbf {\bibinfo {volume}
  {69}},\ \bibinfo {pages} {392--401} (\bibinfo {year} {2019})}\BibitemShut
  {NoStop}%
\bibitem [{\citenamefont {LeCun}\ \emph {et~al.}(1995)\citenamefont {LeCun},
  \citenamefont {Jackel}, \citenamefont {Bottou}, \citenamefont {Cortes},
  \citenamefont {Denker}, \citenamefont {Drucker}, \citenamefont {Guyon},
  \citenamefont {Muller}, \citenamefont {Sackinger}, \citenamefont {Simard}
  \emph {et~al.}}]{lecun1995learning}%
  \BibitemOpen
  \bibfield  {author} {\bibinfo {author} {\bibfnamefont {Yann}\ \bibnamefont
  {LeCun}}, \bibinfo {author} {\bibfnamefont {Lawrence~D}\ \bibnamefont
  {Jackel}}, \bibinfo {author} {\bibfnamefont {L{\'e}on}\ \bibnamefont
  {Bottou}}, \bibinfo {author} {\bibfnamefont {Corinna}\ \bibnamefont
  {Cortes}}, \bibinfo {author} {\bibfnamefont {John~S}\ \bibnamefont {Denker}},
  \bibinfo {author} {\bibfnamefont {Harris}\ \bibnamefont {Drucker}}, \bibinfo
  {author} {\bibfnamefont {Isabelle}\ \bibnamefont {Guyon}}, \bibinfo {author}
  {\bibfnamefont {Urs~A}\ \bibnamefont {Muller}}, \bibinfo {author}
  {\bibfnamefont {Eduard}\ \bibnamefont {Sackinger}}, \bibinfo {author}
  {\bibfnamefont {Patrice}\ \bibnamefont {Simard}},  \emph {et~al.},\
  }\bibfield  {title} {\enquote {\bibinfo {title} {Learning algorithms for
  classification: A comparison on handwritten digit recognition},}\ }\href@noop
  {} {\bibfield  {journal} {\bibinfo  {journal} {Neural networks: the
  statistical mechanics perspective}\ }\textbf {\bibinfo {volume} {261}},\
  \bibinfo {pages} {2} (\bibinfo {year} {1995})}\BibitemShut {NoStop}%
\bibitem [{mni()}]{mnist}%
  \BibitemOpen
  \href@noop {} {}\bibinfo {howpublished}
  {\url{http://yann.lecun.com/exdb/mnist/}}\BibitemShut {NoStop}%
\bibitem [{\citenamefont {Kinzel}(1985)}]{kinzel1985phase}%
  \BibitemOpen
  \bibfield  {author} {\bibinfo {author} {\bibfnamefont {W}~\bibnamefont
  {Kinzel}},\ }\bibfield  {title} {\enquote {\bibinfo {title} {Phase
  transitions of cellular automata},}\ }\href@noop {} {\bibfield  {journal}
  {\bibinfo  {journal} {Zeitschrift f{\"u}r Physik B Condensed Matter}\
  }\textbf {\bibinfo {volume} {58}},\ \bibinfo {pages} {229--244} (\bibinfo
  {year} {1985})}\BibitemShut {NoStop}%
\bibitem [{\citenamefont {Binney}\ \emph {et~al.}(1992)\citenamefont {Binney},
  \citenamefont {Dowrick}, \citenamefont {Fisher},\ and\ \citenamefont
  {Newman}}]{binney1992theory}%
  \BibitemOpen
  \bibfield  {author} {\bibinfo {author} {\bibfnamefont {James~J}\ \bibnamefont
  {Binney}}, \bibinfo {author} {\bibfnamefont {NJ}~\bibnamefont {Dowrick}},
  \bibinfo {author} {\bibfnamefont {AJ}~\bibnamefont {Fisher}}, \ and\ \bibinfo
  {author} {\bibfnamefont {M}~\bibnamefont {Newman}},\ }\href@noop {} {\emph
  {\bibinfo {title} {The theory of critical phenomena: an introduction to the
  renormalization group}}}\ (\bibinfo  {publisher} {Oxford University Press,
  Inc.},\ \bibinfo {year} {1992})\BibitemShut {NoStop}%
\bibitem [{\citenamefont {Hedges}\ \emph {et~al.}(2009)\citenamefont {Hedges},
  \citenamefont {Jack}, \citenamefont {Garrahan},\ and\ \citenamefont
  {Chandler}}]{hedges2009dynamic}%
  \BibitemOpen
  \bibfield  {author} {\bibinfo {author} {\bibfnamefont {Lester~O}\
  \bibnamefont {Hedges}}, \bibinfo {author} {\bibfnamefont {Robert~L}\
  \bibnamefont {Jack}}, \bibinfo {author} {\bibfnamefont {Juan~P}\ \bibnamefont
  {Garrahan}}, \ and\ \bibinfo {author} {\bibfnamefont {David}\ \bibnamefont
  {Chandler}},\ }\bibfield  {title} {\enquote {\bibinfo {title} {Dynamic
  order-disorder in atomistic models of structural glass formers},}\
  }\href@noop {} {\bibfield  {journal} {\bibinfo  {journal} {Science}\ }\textbf
  {\bibinfo {volume} {323}},\ \bibinfo {pages} {1309--1313} (\bibinfo {year}
  {2009})}\BibitemShut {NoStop}%
\bibitem [{\citenamefont {Meshkov}(1997)}]{meshkov1997low}%
  \BibitemOpen
  \bibfield  {author} {\bibinfo {author} {\bibfnamefont {SV}~\bibnamefont
  {Meshkov}},\ }\bibfield  {title} {\enquote {\bibinfo {title} {Low-frequency
  dynamics of {L}ennard-{J}ones glasses},}\ }\href@noop {} {\bibfield
  {journal} {\bibinfo  {journal} {Physical Review B}\ }\textbf {\bibinfo
  {volume} {55}},\ \bibinfo {pages} {12113} (\bibinfo {year}
  {1997})}\BibitemShut {NoStop}%
\bibitem [{\citenamefont {Paugam-Moisy}\ and\ \citenamefont
  {Bohte}(2012)}]{paugam2012computing}%
  \BibitemOpen
  \bibfield  {author} {\bibinfo {author} {\bibfnamefont {H{\'e}lene}\
  \bibnamefont {Paugam-Moisy}}\ and\ \bibinfo {author} {\bibfnamefont
  {Sander~M}\ \bibnamefont {Bohte}},\ }\bibfield  {title} {\enquote {\bibinfo
  {title} {Computing with spiking neuron networks.}}\ }\href@noop {} {\bibfield
   {journal} {\bibinfo  {journal} {Handbook of natural computing}\ }\textbf
  {\bibinfo {volume} {1}},\ \bibinfo {pages} {1--47} (\bibinfo {year}
  {2012})}\BibitemShut {NoStop}%
\bibitem [{\citenamefont {Gerstner}\ and\ \citenamefont
  {Kistler}(2002)}]{gerstner2002spiking}%
  \BibitemOpen
  \bibfield  {author} {\bibinfo {author} {\bibfnamefont {Wulfram}\ \bibnamefont
  {Gerstner}}\ and\ \bibinfo {author} {\bibfnamefont {Werner~M}\ \bibnamefont
  {Kistler}},\ }\href@noop {} {\emph {\bibinfo {title} {Spiking neuron models:
  Single neurons, populations, plasticity}}}\ (\bibinfo  {publisher} {Cambridge
  University Press},\ \bibinfo {year} {2002})\BibitemShut {NoStop}%
\bibitem [{\citenamefont {Jiang}\ and\ \citenamefont
  {Wu}(2002)}]{jiang2002cellular}%
  \BibitemOpen
  \bibfield  {author} {\bibinfo {author} {\bibfnamefont {Rui}\ \bibnamefont
  {Jiang}}\ and\ \bibinfo {author} {\bibfnamefont {Qing-Song}\ \bibnamefont
  {Wu}},\ }\bibfield  {title} {\enquote {\bibinfo {title} {Cellular automata
  models for synchronized traffic flow},}\ }\href@noop {} {\bibfield  {journal}
  {\bibinfo  {journal} {Journal of Physics A: Mathematical and General}\
  }\textbf {\bibinfo {volume} {36}},\ \bibinfo {pages} {381} (\bibinfo {year}
  {2002})}\BibitemShut {NoStop}%
\bibitem [{\citenamefont {Wolf}(1999)}]{wolf1999cellular}%
  \BibitemOpen
  \bibfield  {author} {\bibinfo {author} {\bibfnamefont {Dietrich~E}\
  \bibnamefont {Wolf}},\ }\bibfield  {title} {\enquote {\bibinfo {title}
  {Cellular automata for traffic simulations},}\ }\href@noop {} {\bibfield
  {journal} {\bibinfo  {journal} {Physica A: Statistical Mechanics and its
  Applications}\ }\textbf {\bibinfo {volume} {263}},\ \bibinfo {pages}
  {438--451} (\bibinfo {year} {1999})}\BibitemShut {NoStop}%
\bibitem [{\citenamefont {Neto}\ \emph {et~al.}(2011)\citenamefont {Neto},
  \citenamefont {Lyra},\ and\ \citenamefont {Da~Silva}}]{neto2011phase}%
  \BibitemOpen
  \bibfield  {author} {\bibinfo {author} {\bibfnamefont {JPL}\ \bibnamefont
  {Neto}}, \bibinfo {author} {\bibfnamefont {ML}~\bibnamefont {Lyra}}, \ and\
  \bibinfo {author} {\bibfnamefont {CR}~\bibnamefont {Da~Silva}},\ }\bibfield
  {title} {\enquote {\bibinfo {title} {Phase coexistence induced by a defensive
  reaction in a cellular automaton traffic flow model},}\ }\href@noop {}
  {\bibfield  {journal} {\bibinfo  {journal} {Physica A: Statistical Mechanics
  and its Applications}\ }\textbf {\bibinfo {volume} {390}},\ \bibinfo {pages}
  {3558--3565} (\bibinfo {year} {2011})}\BibitemShut {NoStop}%
\bibitem [{\citenamefont {DÕSouza}(2005)}]{d2005coexisting}%
  \BibitemOpen
  \bibfield  {author} {\bibinfo {author} {\bibfnamefont {Raissa~M}\
  \bibnamefont {DÕSouza}},\ }\bibfield  {title} {\enquote {\bibinfo {title}
  {Coexisting phases and lattice dependence of a cellular automaton model for
  traffic flow},}\ }\href@noop {} {\bibfield  {journal} {\bibinfo  {journal}
  {Physical Review E}\ }\textbf {\bibinfo {volume} {71}},\ \bibinfo {pages}
  {066112} (\bibinfo {year} {2005})}\BibitemShut {NoStop}%
\bibitem [{\citenamefont {Schadschneider}\ \emph {et~al.}(1999)\citenamefont
  {Schadschneider}, \citenamefont {Chowdhury}, \citenamefont {Brockfeld},
  \citenamefont {Klauck}, \citenamefont {Santen},\ and\ \citenamefont
  {Zittartz}}]{schadschneider1999new}%
  \BibitemOpen
  \bibfield  {author} {\bibinfo {author} {\bibfnamefont {A}~\bibnamefont
  {Schadschneider}}, \bibinfo {author} {\bibfnamefont {D}~\bibnamefont
  {Chowdhury}}, \bibinfo {author} {\bibfnamefont {E}~\bibnamefont {Brockfeld}},
  \bibinfo {author} {\bibfnamefont {K}~\bibnamefont {Klauck}}, \bibinfo
  {author} {\bibfnamefont {L}~\bibnamefont {Santen}}, \ and\ \bibinfo {author}
  {\bibfnamefont {J}~\bibnamefont {Zittartz}},\ }\bibfield  {title} {\enquote
  {\bibinfo {title} {A new cellular automata model for city traffic},}\
  }\href@noop {} {\bibfield  {journal} {\bibinfo  {journal} {arXiv preprint
  cond-mat/9911312}\ } (\bibinfo {year} {1999})}\BibitemShut {NoStop}%
\bibitem [{\citenamefont {Yukawa}\ \emph {et~al.}(1994)\citenamefont {Yukawa},
  \citenamefont {Kikuchi},\ and\ \citenamefont {Tadaki}}]{yukawa1994dynamical}%
  \BibitemOpen
  \bibfield  {author} {\bibinfo {author} {\bibfnamefont {Satoshi}\ \bibnamefont
  {Yukawa}}, \bibinfo {author} {\bibfnamefont {Macoto}\ \bibnamefont
  {Kikuchi}}, \ and\ \bibinfo {author} {\bibfnamefont {Shin-ichi}\ \bibnamefont
  {Tadaki}},\ }\bibfield  {title} {\enquote {\bibinfo {title} {Dynamical phase
  transition in one dimensional traffic flow model with blockage},}\
  }\href@noop {} {\bibfield  {journal} {\bibinfo  {journal} {Journal of the
  Physical Society of Japan}\ }\textbf {\bibinfo {volume} {63}},\ \bibinfo
  {pages} {3609--3618} (\bibinfo {year} {1994})}\BibitemShut {NoStop}%
\bibitem [{\citenamefont {Li}\ \emph {et~al.}(1990)\citenamefont {Li},
  \citenamefont {Packard},\ and\ \citenamefont {Langton}}]{li1990transition}%
  \BibitemOpen
  \bibfield  {author} {\bibinfo {author} {\bibfnamefont {Wentian}\ \bibnamefont
  {Li}}, \bibinfo {author} {\bibfnamefont {Norman~H}\ \bibnamefont {Packard}},
  \ and\ \bibinfo {author} {\bibfnamefont {Chris~G}\ \bibnamefont {Langton}},\
  }\bibfield  {title} {\enquote {\bibinfo {title} {Transition phenomena in
  cellular automata rule space},}\ }\href@noop {} {\bibfield  {journal}
  {\bibinfo  {journal} {Physica D: Nonlinear Phenomena}\ }\textbf {\bibinfo
  {volume} {45}},\ \bibinfo {pages} {77--94} (\bibinfo {year}
  {1990})}\BibitemShut {NoStop}%
\bibitem [{\citenamefont {Wootters}\ and\ \citenamefont
  {Langton}(1990)}]{wootters1990there}%
  \BibitemOpen
  \bibfield  {author} {\bibinfo {author} {\bibfnamefont {William~K}\
  \bibnamefont {Wootters}}\ and\ \bibinfo {author} {\bibfnamefont {Chris~G}\
  \bibnamefont {Langton}},\ }\bibfield  {title} {\enquote {\bibinfo {title} {Is
  there a sharp phase transition for deterministic cellular automata?}}\
  }\href@noop {} {\bibfield  {journal} {\bibinfo  {journal} {Physica D:
  Nonlinear Phenomena}\ }\textbf {\bibinfo {volume} {45}},\ \bibinfo {pages}
  {95--104} (\bibinfo {year} {1990})}\BibitemShut {NoStop}%
\bibitem [{\citenamefont {Klymko}\ \emph {et~al.}(2017)\citenamefont {Klymko},
  \citenamefont {Garrahan},\ and\ \citenamefont
  {Whitelam}}]{klymko2017similarity}%
  \BibitemOpen
  \bibfield  {author} {\bibinfo {author} {\bibfnamefont {Katherine}\
  \bibnamefont {Klymko}}, \bibinfo {author} {\bibfnamefont {Juan~P}\
  \bibnamefont {Garrahan}}, \ and\ \bibinfo {author} {\bibfnamefont {Stephen}\
  \bibnamefont {Whitelam}},\ }\bibfield  {title} {\enquote {\bibinfo {title}
  {Similarity of ensembles of trajectories of reversible and irreversible
  growth processes},}\ }\href@noop {} {\bibfield  {journal} {\bibinfo
  {journal} {Physical Review E}\ }\textbf {\bibinfo {volume} {96}},\ \bibinfo
  {pages} {042126} (\bibinfo {year} {2017})}\BibitemShut {NoStop}%
\bibitem [{\citenamefont {Packard}\ and\ \citenamefont
  {Wolfram}(1985)}]{packard1985two}%
  \BibitemOpen
  \bibfield  {author} {\bibinfo {author} {\bibfnamefont {Norman~H}\
  \bibnamefont {Packard}}\ and\ \bibinfo {author} {\bibfnamefont {Stephen}\
  \bibnamefont {Wolfram}},\ }\bibfield  {title} {\enquote {\bibinfo {title}
  {Two-dimensional cellular automata},}\ }\href@noop {} {\bibfield  {journal}
  {\bibinfo  {journal} {Journal of Statistical physics}\ }\textbf {\bibinfo
  {volume} {38}},\ \bibinfo {pages} {901--946} (\bibinfo {year}
  {1985})}\BibitemShut {NoStop}%
\bibitem [{Note1()}]{Note1}%
  \BibitemOpen
  \bibinfo {note} {Also known as random-mutation hill climbing~\cite
  {mitchell1993will,mitchell1998introduction}.}\BibitemShut {Stop}%
\bibitem [{Note2()}]{Note2}%
  \BibitemOpen
  \bibinfo {note} {This approach is similar to those that use genetic
  algorithms to search for automaton rules that perform particular
  computations~\cite
  {suzudo2004searching,mitchell1996evolving,oliveira2009some,chavoya2006using},
  except that the present search scheme applies mutations and the Metropolis
  acceptance criterion to a single individual, rather than applying genetic
  operations to a population of individuals. For continuous variables and small
  mutations, the present procedure constitutes noisy clipped gradient descent
  on the loss surface~\cite {whitelam2021correspondence}.}\BibitemShut {Stop}%
\bibitem [{\citenamefont {Gers}\ \emph {et~al.}(1997)\citenamefont {Gers},
  \citenamefont {Garis},\ and\ \citenamefont {Korkin}}]{gers1997codi}%
  \BibitemOpen
  \bibfield  {author} {\bibinfo {author} {\bibfnamefont {Felix}\ \bibnamefont
  {Gers}}, \bibinfo {author} {\bibfnamefont {Hugo~de}\ \bibnamefont {Garis}}, \
  and\ \bibinfo {author} {\bibfnamefont {Michael}\ \bibnamefont {Korkin}},\
  }\bibfield  {title} {\enquote {\bibinfo {title} {Codi-1bit: A simplified
  cellular automata based neuron model},}\ }in\ \href@noop {} {\emph {\bibinfo
  {booktitle} {European Conference on Artificial Evolution}}}\ (\bibinfo
  {organization} {Springer},\ \bibinfo {year} {1997})\ pp.\ \bibinfo {pages}
  {315--333}\BibitemShut {NoStop}%
\bibitem [{\citenamefont {Whitelam}\ and\ \citenamefont
  {Jacobson}(2021)}]{whitelam2021varied}%
  \BibitemOpen
  \bibfield  {author} {\bibinfo {author} {\bibfnamefont {Stephen}\ \bibnamefont
  {Whitelam}}\ and\ \bibinfo {author} {\bibfnamefont {Daniel}\ \bibnamefont
  {Jacobson}},\ }\bibfield  {title} {\enquote {\bibinfo {title} {Varied
  phenomenology of models displaying dynamical large-deviation
  singularities},}\ }\href@noop {} {\bibfield  {journal} {\bibinfo  {journal}
  {Physical Review E}\ }\textbf {\bibinfo {volume} {103}},\ \bibinfo {pages}
  {032152} (\bibinfo {year} {2021})}\BibitemShut {NoStop}%
\bibitem [{\citenamefont {Schmidhuber}(1997)}]{schmidhuber1997discovering}%
  \BibitemOpen
  \bibfield  {author} {\bibinfo {author} {\bibfnamefont {J{\"u}rgen}\
  \bibnamefont {Schmidhuber}},\ }\bibfield  {title} {\enquote {\bibinfo {title}
  {Discovering neural nets with low kolmogorov complexity and high
  generalization capability},}\ }\href@noop {} {\bibfield  {journal} {\bibinfo
  {journal} {Neural Networks}\ }\textbf {\bibinfo {volume} {10}},\ \bibinfo
  {pages} {857--873} (\bibinfo {year} {1997})}\BibitemShut {NoStop}%
\bibitem [{\citenamefont {Schmidhuber}(2015)}]{schmidhuber2015deep}%
  \BibitemOpen
  \bibfield  {author} {\bibinfo {author} {\bibfnamefont {J{\"u}rgen}\
  \bibnamefont {Schmidhuber}},\ }\bibfield  {title} {\enquote {\bibinfo {title}
  {Deep learning in neural networks: An overview},}\ }\href@noop {} {\bibfield
  {journal} {\bibinfo  {journal} {Neural networks}\ }\textbf {\bibinfo {volume}
  {61}},\ \bibinfo {pages} {85--117} (\bibinfo {year} {2015})}\BibitemShut
  {NoStop}%
\bibitem [{\citenamefont {Goodfellow}\ \emph {et~al.}(2016)\citenamefont
  {Goodfellow}, \citenamefont {Bengio},\ and\ \citenamefont
  {Courville}}]{goodfellow2016deep}%
  \BibitemOpen
  \bibfield  {author} {\bibinfo {author} {\bibfnamefont {Ian}\ \bibnamefont
  {Goodfellow}}, \bibinfo {author} {\bibfnamefont {Yoshua}\ \bibnamefont
  {Bengio}}, \ and\ \bibinfo {author} {\bibfnamefont {Aaron}\ \bibnamefont
  {Courville}},\ }\href@noop {} {\emph {\bibinfo {title} {Deep learning}}}\
  (\bibinfo  {publisher} {MIT press},\ \bibinfo {year} {2016})\BibitemShut
  {NoStop}%
\bibitem [{Note3()}]{Note3}%
  \BibitemOpen
  \bibinfo {note} {Times of snapshots for each row are as follows. 0:
  $10,38,200$; 1: $3,9,200$; 2: $3,13,200$; 7: $5,10,200$.}\BibitemShut {Stop}%
\bibitem [{\citenamefont {Tanaka}\ \emph {et~al.}(2019)\citenamefont {Tanaka},
  \citenamefont {Yamane}, \citenamefont {H{\'e}roux}, \citenamefont {Nakane},
  \citenamefont {Kanazawa}, \citenamefont {Takeda}, \citenamefont {Numata},
  \citenamefont {Nakano},\ and\ \citenamefont {Hirose}}]{tanaka2019recent}%
  \BibitemOpen
  \bibfield  {author} {\bibinfo {author} {\bibfnamefont {Gouhei}\ \bibnamefont
  {Tanaka}}, \bibinfo {author} {\bibfnamefont {Toshiyuki}\ \bibnamefont
  {Yamane}}, \bibinfo {author} {\bibfnamefont {Jean~Benoit}\ \bibnamefont
  {H{\'e}roux}}, \bibinfo {author} {\bibfnamefont {Ryosho}\ \bibnamefont
  {Nakane}}, \bibinfo {author} {\bibfnamefont {Naoki}\ \bibnamefont
  {Kanazawa}}, \bibinfo {author} {\bibfnamefont {Seiji}\ \bibnamefont
  {Takeda}}, \bibinfo {author} {\bibfnamefont {Hidetoshi}\ \bibnamefont
  {Numata}}, \bibinfo {author} {\bibfnamefont {Daiju}\ \bibnamefont {Nakano}},
  \ and\ \bibinfo {author} {\bibfnamefont {Akira}\ \bibnamefont {Hirose}},\
  }\bibfield  {title} {\enquote {\bibinfo {title} {Recent advances in physical
  reservoir computing: A review},}\ }\href@noop {} {\bibfield  {journal}
  {\bibinfo  {journal} {Neural Networks}\ }\textbf {\bibinfo {volume} {115}},\
  \bibinfo {pages} {100--123} (\bibinfo {year} {2019})}\BibitemShut {NoStop}%
\bibitem [{\citenamefont {Kingma}\ and\ \citenamefont
  {Ba}(2014)}]{kingma2014adam}%
  \BibitemOpen
  \bibfield  {author} {\bibinfo {author} {\bibfnamefont {Diederik~P}\
  \bibnamefont {Kingma}}\ and\ \bibinfo {author} {\bibfnamefont {Jimmy}\
  \bibnamefont {Ba}},\ }\bibfield  {title} {\enquote {\bibinfo {title} {Adam: A
  method for stochastic optimization},}\ }\href@noop {} {\bibfield  {journal}
  {\bibinfo  {journal} {arXiv preprint arXiv:1412.6980}\ } (\bibinfo {year}
  {2014})}\BibitemShut {NoStop}%
\bibitem [{\citenamefont {Paszke}\ \emph {et~al.}(2019)\citenamefont {Paszke},
  \citenamefont {Gross}, \citenamefont {Massa}, \citenamefont {Lerer},
  \citenamefont {Bradbury}, \citenamefont {Chanan}, \citenamefont {Killeen},
  \citenamefont {Lin}, \citenamefont {Gimelshein}, \citenamefont {Antiga} \emph
  {et~al.}}]{paszke2019pytorch}%
  \BibitemOpen
  \bibfield  {author} {\bibinfo {author} {\bibfnamefont {Adam}\ \bibnamefont
  {Paszke}}, \bibinfo {author} {\bibfnamefont {Sam}\ \bibnamefont {Gross}},
  \bibinfo {author} {\bibfnamefont {Francisco}\ \bibnamefont {Massa}}, \bibinfo
  {author} {\bibfnamefont {Adam}\ \bibnamefont {Lerer}}, \bibinfo {author}
  {\bibfnamefont {James}\ \bibnamefont {Bradbury}}, \bibinfo {author}
  {\bibfnamefont {Gregory}\ \bibnamefont {Chanan}}, \bibinfo {author}
  {\bibfnamefont {Trevor}\ \bibnamefont {Killeen}}, \bibinfo {author}
  {\bibfnamefont {Zeming}\ \bibnamefont {Lin}}, \bibinfo {author}
  {\bibfnamefont {Natalia}\ \bibnamefont {Gimelshein}}, \bibinfo {author}
  {\bibfnamefont {Luca}\ \bibnamefont {Antiga}},  \emph {et~al.},\ }\bibfield
  {title} {\enquote {\bibinfo {title} {Pytorch: An imperative style,
  high-performance deep learning library},}\ }\href@noop {} {\bibfield
  {journal} {\bibinfo  {journal} {Advances in neural information processing
  systems}\ }\textbf {\bibinfo {volume} {32}} (\bibinfo {year}
  {2019})}\BibitemShut {NoStop}%
\bibitem [{\citenamefont {Mitchell}\ \emph {et~al.}(1994)\citenamefont
  {Mitchell}, \citenamefont {Crutchfield},\ and\ \citenamefont
  {Hraber}}]{mitchell1994dynamics}%
  \BibitemOpen
  \bibfield  {author} {\bibinfo {author} {\bibfnamefont {Melanie}\ \bibnamefont
  {Mitchell}}, \bibinfo {author} {\bibfnamefont {James~P}\ \bibnamefont
  {Crutchfield}}, \ and\ \bibinfo {author} {\bibfnamefont {Peter~T}\
  \bibnamefont {Hraber}},\ }\bibfield  {title} {\enquote {\bibinfo {title}
  {Dynamics, computation, and the edge of chaos": A re-examination},}\ }in\
  \href@noop {} {\emph {\bibinfo {booktitle} {Santa Fe Institute Studies in the
  Sciences of Complexity}}},\ Vol.~\bibinfo {volume} {19}\ (\bibinfo
  {organization} {Addison-Wesley Publishing Co.},\ \bibinfo {year} {1994})\
  pp.\ \bibinfo {pages} {497--497}\BibitemShut {NoStop}%
\bibitem [{\citenamefont {Langton}(1990)}]{langton1990computation}%
  \BibitemOpen
  \bibfield  {author} {\bibinfo {author} {\bibfnamefont {Chris~G}\ \bibnamefont
  {Langton}},\ }\bibfield  {title} {\enquote {\bibinfo {title} {Computation at
  the edge of chaos: Phase transitions and emergent computation},}\ }\href@noop
  {} {\bibfield  {journal} {\bibinfo  {journal} {Physica D: nonlinear
  phenomena}\ }\textbf {\bibinfo {volume} {42}},\ \bibinfo {pages} {12--37}
  (\bibinfo {year} {1990})}\BibitemShut {NoStop}%
\bibitem [{\citenamefont {Wolfram}(1985)}]{wolfram1985twenty}%
  \BibitemOpen
  \bibfield  {author} {\bibinfo {author} {\bibfnamefont {Stephen}\ \bibnamefont
  {Wolfram}},\ }\bibfield  {title} {\enquote {\bibinfo {title} {Twenty problems
  in the theory of cellular automata},}\ }\href@noop {} {\bibfield  {journal}
  {\bibinfo  {journal} {Physica Scripta}\ }\textbf {\bibinfo {volume} {1985}},\
  \bibinfo {pages} {170} (\bibinfo {year} {1985})}\BibitemShut {NoStop}%
\bibitem [{\citenamefont {Mitchell}\ \emph {et~al.}(1993)\citenamefont
  {Mitchell}, \citenamefont {Holland},\ and\ \citenamefont
  {Forrest}}]{mitchell1993will}%
  \BibitemOpen
  \bibfield  {author} {\bibinfo {author} {\bibfnamefont {Melanie}\ \bibnamefont
  {Mitchell}}, \bibinfo {author} {\bibfnamefont {John}\ \bibnamefont
  {Holland}}, \ and\ \bibinfo {author} {\bibfnamefont {Stephanie}\ \bibnamefont
  {Forrest}},\ }\bibfield  {title} {\enquote {\bibinfo {title} {When will a
  genetic algorithm outperform hill climbing},}\ }\href@noop {} {\bibfield
  {journal} {\bibinfo  {journal} {Advances in neural information processing
  systems}\ }\textbf {\bibinfo {volume} {6}} (\bibinfo {year}
  {1993})}\BibitemShut {NoStop}%
\bibitem [{\citenamefont {Mitchell}(1998)}]{mitchell1998introduction}%
  \BibitemOpen
  \bibfield  {author} {\bibinfo {author} {\bibfnamefont {Melanie}\ \bibnamefont
  {Mitchell}},\ }\href@noop {} {\emph {\bibinfo {title} {An introduction to
  genetic algorithms}}}\ (\bibinfo  {publisher} {MIT press},\ \bibinfo {year}
  {1998})\BibitemShut {NoStop}%
\bibitem [{\citenamefont {Suzudo}(2004)}]{suzudo2004searching}%
  \BibitemOpen
  \bibfield  {author} {\bibinfo {author} {\bibfnamefont {Tomoaki}\ \bibnamefont
  {Suzudo}},\ }\bibfield  {title} {\enquote {\bibinfo {title} {Searching for
  pattern-forming asynchronous cellular automata--an evolutionary approach},}\
  }in\ \href@noop {} {\emph {\bibinfo {booktitle} {International Conference on
  Cellular Automata}}}\ (\bibinfo {organization} {Springer},\ \bibinfo {year}
  {2004})\ pp.\ \bibinfo {pages} {151--160}\BibitemShut {NoStop}%
\bibitem [{\citenamefont {Mitchell}\ \emph {et~al.}(1996)\citenamefont
  {Mitchell}, \citenamefont {Crutchfield}, \citenamefont {Das} \emph
  {et~al.}}]{mitchell1996evolving}%
  \BibitemOpen
  \bibfield  {author} {\bibinfo {author} {\bibfnamefont {Melanie}\ \bibnamefont
  {Mitchell}}, \bibinfo {author} {\bibfnamefont {James~P}\ \bibnamefont
  {Crutchfield}}, \bibinfo {author} {\bibfnamefont {Rajarshi}\ \bibnamefont
  {Das}},  \emph {et~al.},\ }\bibfield  {title} {\enquote {\bibinfo {title}
  {Evolving cellular automata with genetic algorithms: A review of recent
  work},}\ }in\ \href@noop {} {\emph {\bibinfo {booktitle} {Proceedings of the
  First International Conference on Evolutionary Computation and Its
  Applications (EvCA96)}}},\ Vol.~\bibinfo {volume} {8}\ (\bibinfo
  {organization} {Moscow},\ \bibinfo {year} {1996})\BibitemShut {NoStop}%
\bibitem [{\citenamefont {Oliveira}\ \emph {et~al.}(2009)\citenamefont
  {Oliveira}, \citenamefont {Martins}, \citenamefont {de~Carvalho},\ and\
  \citenamefont {Fynn}}]{oliveira2009some}%
  \BibitemOpen
  \bibfield  {author} {\bibinfo {author} {\bibfnamefont {Gina~MB}\ \bibnamefont
  {Oliveira}}, \bibinfo {author} {\bibfnamefont {Luiz~GA}\ \bibnamefont
  {Martins}}, \bibinfo {author} {\bibfnamefont {Laura~B}\ \bibnamefont
  {de~Carvalho}}, \ and\ \bibinfo {author} {\bibfnamefont {Enrique}\
  \bibnamefont {Fynn}},\ }\bibfield  {title} {\enquote {\bibinfo {title} {Some
  investigations about synchronization and density classification tasks in
  one-dimensional and two-dimensional cellular automata rule spaces},}\
  }\href@noop {} {\bibfield  {journal} {\bibinfo  {journal} {Electronic Notes
  in Theoretical Computer Science}\ }\textbf {\bibinfo {volume} {252}},\
  \bibinfo {pages} {121--142} (\bibinfo {year} {2009})}\BibitemShut {NoStop}%
\bibitem [{\citenamefont {Chavoya}\ and\ \citenamefont
  {Duthen}(2006)}]{chavoya2006using}%
  \BibitemOpen
  \bibfield  {author} {\bibinfo {author} {\bibfnamefont {Arturo}\ \bibnamefont
  {Chavoya}}\ and\ \bibinfo {author} {\bibfnamefont {Yves}\ \bibnamefont
  {Duthen}},\ }\bibfield  {title} {\enquote {\bibinfo {title} {Using a genetic
  algorithm to evolve cellular automata for 2d/3d computational development},}\
  }in\ \href@noop {} {\emph {\bibinfo {booktitle} {Proceedings of the 8th
  annual conference on Genetic and evolutionary computation}}}\ (\bibinfo
  {year} {2006})\ pp.\ \bibinfo {pages} {231--232}\BibitemShut {NoStop}%
\bibitem [{\citenamefont {Whitelam}\ \emph {et~al.}(2021)\citenamefont
  {Whitelam}, \citenamefont {Selin}, \citenamefont {Park},\ and\ \citenamefont
  {Tamblyn}}]{whitelam2021correspondence}%
  \BibitemOpen
  \bibfield  {author} {\bibinfo {author} {\bibfnamefont {Stephen}\ \bibnamefont
  {Whitelam}}, \bibinfo {author} {\bibfnamefont {Viktor}\ \bibnamefont
  {Selin}}, \bibinfo {author} {\bibfnamefont {Sang-Won}\ \bibnamefont {Park}},
  \ and\ \bibinfo {author} {\bibfnamefont {Isaac}\ \bibnamefont {Tamblyn}},\
  }\bibfield  {title} {\enquote {\bibinfo {title} {Correspondence between
  neuroevolution and gradient descent},}\ }\href@noop {} {\bibfield  {journal}
  {\bibinfo  {journal} {Nature communications}\ }\textbf {\bibinfo {volume}
  {12}},\ \bibinfo {pages} {1--10} (\bibinfo {year} {2021})}\BibitemShut
  {NoStop}%
\end{thebibliography}

%

\appendix
\section{Example automata}
\label{examples}

In this section we specify three automata that learned, under variable-depth search, to recognize three different digit classes via the dynamical phase coexistence mechanism described in the text. Entries in the rule table are numbered sequentially from 0 to 511, according to \eqq{enviro} and the associated description. 

\begin{itemize}

\item Class 0: depth $T=512$, rule table ${\cal R}=$

 {\small \tt 0 0 0 0 0 0 0 1 1 0 1 0 0 1 1 1 0 1 1 1 1 0 0 0 1 0 1 0 1 0 0 1 1 0 1 1 0 0 0 1 0 1 1 1 0 0 0 1 0 0 1 1 0 0 0 1 0 0 0 1 0 1 0 1 0 0 0 1 0 0 1 1 0 0 1 1 1 0 1 0 1 0 1 1 0 0 0 0 0 1 0 1 0 1 0 1 1 1 1 1 1 0 1 0 1 0 1 1 1 1 1 1 0 0 0 0 0 0 0 0 0 1 0 1 1 1 0 1 0 0 0 0 0 0 1 0 1 1 1 1 1 1 1 1 1 0 1 1 0 0 1 0 0 0 0 0 0 1 1 1 1 1 1 1 0 1 1 1 1 1 1 1 0 1 1 1 1 1 0 1 0 0 1 0 1 0 0 0 0 1 0 1 0 0 1 1 0 0 0 0 0 0 1 1 1 1 0 1 0 0 0 0 1 0 1 1 0 1 0 1 1 1 1 1 1 1 1 1 0 1 1 1 1 1 0 1 0 1 0 1 1 1 1 0 0 1 1 1 0 0 1 1 0 0 1 1 0 1 0 0 0 0 1 0 0 0 1 0 0 0 0 1 0 0 0 0 0 0 0 0 0 0 0 0 0 0 0 0 0 0 1 0 0 0 0 1 0 1 1 1 0 0 0 0 1 0 0 0 0 1 1 1 1 0 0 1 0 0 0 1 0 0 0 0 0 0 1 0 0 0 1 1 0 0 1 0 1 0 0 1 0 0 0 0 0 0 1 0 0 0 0 1 0 1 1 1 1 0 0 1 0 1 0 1 0 1 0 1 0 0 1 0 0 0 0 1 0 1 0 0 0 0 0 1 1 1 1 1 0 0 0 1 0 0 1 1 0 1 1 0 1 1 0 0 0 0 0 0 1 1 1 1 0 0 0 1 0 0 1 0 0 0 1 1 0 1 1 1 0 0 1 1 0 0 0 1 1 1 1 1 0 0 0 1 0 1 0 1 0 0 1 1 1 1 1 1 0 1 0 1 0 1 1 1 0 1 1 0 1 0 0 0 1 1 1 1 1 1 0 1 0 0 0 1 0 0 0 0 1 0 0 1 0 0 1 1 1 0 1 1 1 1 1 1 1 1 0 1 1 1 1 1}\\

\item Class 1: depth $T=698$, rule table ${\cal R}=$

 {\small \tt 0 0 0 1 0 1 0 1 0 0 0 0 0 0 1 1 0 1 0 0 0 1 0 0 0 1 0 1 0 0 0 0 0 1 1 1 0 1 0 1 0 0 0 1 0 1 1 1 0 0 1 0 0 0 1 1 0 0 0 0 0 1 0 0 0 0 0 0 0 1 0 1 0 0 0 0 1 0 1 1 0 0 0 0 0 0 0 1 0 1 0 1 0 0 1 1 0 1 0 1 0 0 1 1 0 0 1 1 0 1 0 1 0 0 0 0 0 0 1 0 0 0 0 0 0 0 1 0 0 0 1 1 0 1 1 1 0 0 1 0 0 0 1 0 0 0 1 1 1 1 1 1 0 1 0 1 0 1 0 0 0 1 1 1 0 1 1 1 1 0 1 0 0 0 0 1 1 0 1 0 1 0 1 1 0 0 1 0 1 1 0 1 0 0 1 1 1 1 1 1 1 0 0 1 0 0 0 1 1 0 0 1 0 1 0 0 0 1 1 1 0 0 1 0 1 1 1 1 1 1 1 0 0 0 1 0 0 1 0 0 0 0 0 1 1 1 0 1 0 0 0 1 0 0 1 1 1 1 0 1 0 0 0 1 0 1 1 1 0 1 1 1 0 0 1 1 0 0 1 0 0 0 0 1 0 0 0 0 0 1 1 1 0 0 0 1 1 0 0 1 0 1 1 0 0 0 0 0 0 0 1 1 0 0 0 0 0 0 1 0 0 1 0 1 1 1 1 1 0 0 1 0 0 0 1 1 1 0 0 0 0 1 0 0 0 1 1 1 0 1 0 0 1 1 1 1 0 1 0 1 0 0 1 0 0 0 1 0 1 0 0 0 0 1 0 0 1 1 0 1 0 1 1 0 1 1 1 1 0 1 1 1 1 0 1 1 1 0 1 1 0 0 1 1 0 0 1 1 0 1 0 1 1 0 0 0 1 1 1 1 0 1 1 1 1 0 1 0 1 1 0 1 1 0 1 1 0 0 1 1 0 1 0 0 0 0 0 0 1 1 0 1 1 1 1 1 1 0 1 0 1 0 1 1 0 0 0 0 1 1 1 0 0 1 0 1 0 1 0 1 1 1 0 1 1 1 0 1 1 0 1 0 1 1 1 1 0 0 0 1 1 0 1 0 0 0 0 1 1 1 0 0}\\

\item Class 3: depth $T=699$, rule table ${\cal R}=$

{\small \tt 0 0 0 0 0 0 0 0 0 0 0 0 1 1 0 0 1 0 1 1 1 0 1 1 1 1 1 1 1 1 1 1 1 0 1 0 1 1 0 1 0 1 0 0 0 1 1 1 1 1 0 0 0 1 1 1 1 1 1 1 1 0 1 1 1 0 0 0 0 0 1 1 1 1 0 0 0 1 1 1 1 1 0 1 1 0 1 0 0 1 0 1 1 1 1 1 1 0 1 0 0 0 1 1 1 1 0 1 1 1 0 1 0 0 1 1 1 0 0 0 1 1 1 0 1 0 0 1 0 0 0 0 0 0 1 0 1 1 0 0 0 1 1 1 1 1 0 0 1 0 0 0 1 1 0 0 1 0 1 1 0 0 0 1 0 1 0 1 0 1 1 0 0 1 1 1 1 1 0 0 1 1 0 1 1 0 1 0 0 1 0 1 0 0 1 0 0 0 1 0 1 1 0 0 1 0 1 1 0 1 0 0 1 0 0 1 1 0 0 0 0 0 0 1 0 0 0 1 1 1 1 0 1 0 1 0 1 1 0 0 1 0 0 1 1 1 0 1 1 1 1 1 1 1 1 1 0 0 0 0 0 0 1 0 0 0 0 0 0 1 0 0 1 1 0 1 1 0 0 0 1 1 1 0 1 1 0 1 1 1 1 1 1 1 0 1 1 1 0 1 0 1 1 0 1 1 1 0 0 1 1 1 1 1 0 0 1 0 1 1 1 0 1 1 0 1 0 1 0 1 0 0 1 1 1 1 1 0 0 1 1 1 1 1 1 0 0 0 1 0 1 1 1 0 1 1 0 1 1 1 0 1 1 1 1 1 1 1 1 0 1 0 1 1 1 1 1 0 0 0 1 1 0 1 0 0 1 0 1 1 0 1 0 0 0 0 0 1 1 1 1 0 0 1 1 0 0 1 0 1 0 0 1 0 1 1 0 0 0 1 1 1 1 1 1 1 1 0 1 1 1 1 1 1 0 0 0 0 1 1 1 1 0 0 1 1 1 1 0 1 1 0 0 0 1 1 0 1 0 0 1 1 1 1 1 1 0 0 1 0 0 0 1 0 1 0 1 1 0 1 1 1 0 1 0 1 1 0 1 1 0 1 1 1 1 1 1 1 0 1 1 1 0 1 1 1 1 1 1 1 1 1}

\end{itemize}

 \end{document}